\def\year{2022}\relax
\documentclass[letterpaper]{article} 
\usepackage{aaai22}  
\usepackage{times}  
\usepackage{helvet}  
\usepackage{courier}  
\usepackage[hyphens]{url}  
\usepackage{graphicx} 
\usepackage{natbib}  
\usepackage{caption} 
\DeclareCaptionStyle{ruled}{labelfont=normalfont,labelsep=colon,strut=off} 
\usepackage[ruled,vlined]{algorithm2e}
\usepackage{booktabs}
\usepackage{tabularx}
\usepackage{amsfonts}

\graphicspath{{./figures/}}

\title{Prune and Tune Ensembles: Low-Cost Ensemble Learning With Sparse Independent Subnetworks}
\author {
    Tim Whitaker,
    Darrell Whitley
}
\affiliations {
    Department of Computer Science\\
    Colorado State University\\
    Fort Collins, CO 80525 \\
    timothy.whitaker@colostate.edu, whitley@cs.colostate.edu
}

\begin{document}

\maketitle

\begin{abstract}
Ensemble Learning is an effective method for improving generalization in machine learning. However, as state-of-the-art neural networks grow larger, the computational cost associated with training several independent networks becomes expensive. We introduce a fast, low-cost method for creating diverse ensembles of neural networks without needing to train multiple models from scratch. We do this by first training a single parent network. We then create child networks by cloning the parent and dramatically pruning the parameters of each child to create an ensemble of members with unique and diverse topologies. We then briefly train each child network for a small number of epochs, which now converge significantly faster when compared to training from scratch. We explore various ways to maximize diversity in the child networks, including the use of anti-random pruning and one-cycle tuning. This diversity enables ``Prune and Tune" ensembles to achieve results that are competitive with traditional ensembles at a fraction of the training cost. We benchmark our approach against state of the art low-cost ensemble methods and display marked improvement in both accuracy and uncertainty estimation on CIFAR-10 and CIFAR-100.
\end{abstract}

\section{Introduction}

Ensemble learning has long been an active area of research in machine learning \cite{hansen1990neural, krogh1994neural}. Combining the predictions of several models is a simple way to improve generalization on unseen data and ensembles of neural networks have been used to win many high profile machine learning competitions \cite{wolfinger2017stacked}. However, as datasets and neural networks grow larger and more complex, traditional ensemble methods become prohibitively expensive to implement.

Several approaches have been introduced to reduce the large computational costs associated with building ensembles of neural networks. Methods such as pseudo-ensembles, temporal ensembles, and evolutionary ensembles either share network structure or training information among ensemble members in order to reduce the need to train several independent networks from scratch. While these methods are effective at reducing computational cost, they can often be limited in potential ensemble size, member diversity and convergence efficiency. This is a critical problem as ensemble performance increases with the number of well trained and diverse models it contains \cite{bonab2016theoretical, oshiro2012forest}.

We introduce an improved low-cost method for dynamically constructing an ensemble of size $M$. Rather than training several independent networks from scratch, we instead start by training only a single large parent network. We then clone the trained network $M$ times and randomly prune the weights of each clone to create $M$ significantly smaller child networks, each with a unique connective topology. Each child is then fine tuned for a small number of training epochs. We explore several methods to maximize diversity among the generated child networks, including {\em anti-random pruning} \cite{malaiya1995antirandom,shen2008antirandom} and {\em one-cycle tuning} \cite{smith2018superconvergence}.

We call the resulting ensemble a {\em Prune and Tune Ensemble} (PAT Ensemble). Our approach is highly flexible and offers several unique benefits over other low-cost ensemble methods. Since ensemble members are created independently of the parent's training phase, any network architecture, pre-trained or not, can be used as a parent without modification. This allows for very large ensembles as child networks can be dynamically generated with little additional computation. Using pruning methods to create sparse children significantly reduces memory usage and computational cost with no discernible loss in accuracy \cite{blalock2020state}. Our examination of the loss landscapes of child networks show that the combination of anti-random pruning and one-cycle tuning ensures that children converge to unique local optima despite inheriting their parameters from the same network.

We compare Prune and Tune Ensembles to popular low-cost ensemble learning algorithms using several deep neural network architectures on benchmark computer vision datasets. We conduct hyperparameter ablation studies with child sparsity, pruning structures, tuning schedules, and ensemble sizes up to 128 members. We demonstrate that Prune and Tune Ensembles are both highly efficient and highly robust in low training budget (16 epochs) and high training budget (200 epochs) environments. With the experiments introduced in this paper, Prune and Tune Ensembles prove to be more accurate and more diverse than current state-of-the-art low-cost ensemble algorithms, while using significantly fewer parameters.

\begin{figure*}[t]
    \centering
    \includegraphics[width=2\columnwidth]{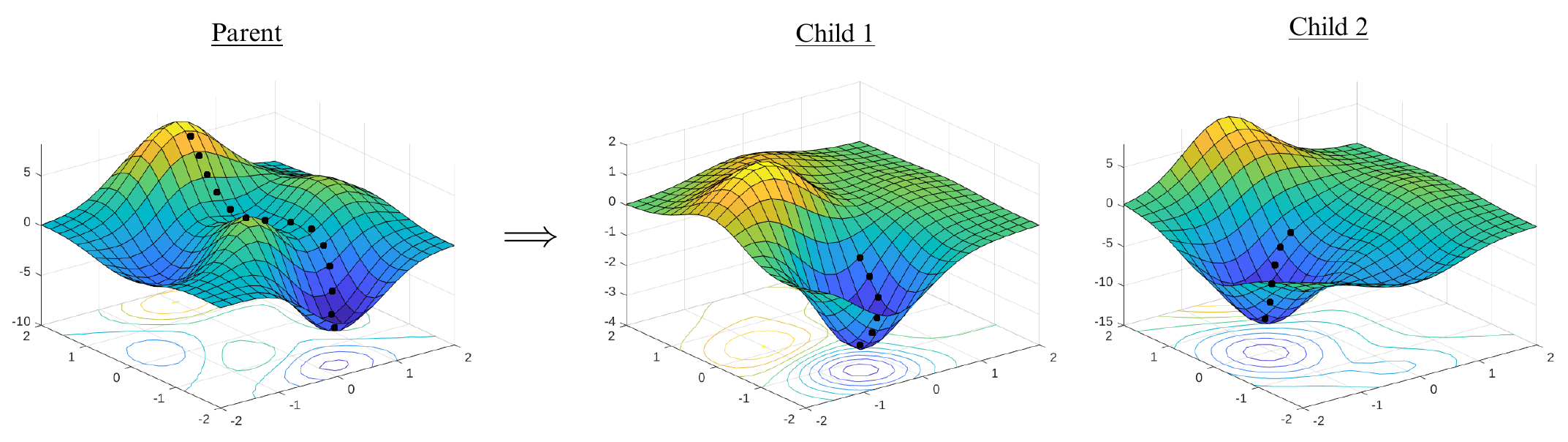}
    \caption{Prune and Tune Ensembles create diverse children via
    transformations that affect the underlying optimization landscapes. Each child explores a unique loss landscape and converges rapidly due to the parameters inherited from the parent.}
    \label{fig:intro-visual}
\end{figure*}

\section{Related Work}

Several approaches have been introduced to reduce the computational cost of constructing ensembles while seeking the associated benefits in generalization. These approaches can be broadly categorized as pseudo-ensembles, temporal ensembles and evolutionary ensembles.

\textbf{Pseudo-Ensembles} involve the differential learning of parameters under the guise of a single monolithic architecture. Dropout is the canonical example where neurons are randomly masked for each mini-batch during training \cite{srivastava2014dropout}. DropConnect masks connections during training instead of neurons \cite{wan2013regularization} and Stochastic Depth Networks mask entire layers of deep residual networks \cite{huang2016stochastic}. TreeNets use multiple independent branches and output heads \cite{lee2015m}. Multi-Input Multi-Output (MIMO) trains a single network on multiple samples simultaneously with distinct input and output heads \cite{havasi2021training}. BatchEnsemble decomposes weight matrices into Hadamard products of a shared weight matrix and a rank-1 matrix for each ensemble member. This allows for simultaneous training of several members on a single forward pass \cite{wen2020batchensemble}. HyperEnsembles vary hyperparameter configurations between members \cite{wenzel2021hyperparameter}. Late Phase Weights is an approach that averages over subsets of weights learned in the late phases of training \cite{vonoswald2021neural}.

Pseudo-Ensembles are memory efficient, but they tend to be less diverse than true ensembles because parameters and network structure are implicitly shared \cite{bachman2014learning}. This lack of diversity can limit the generalization benefits that are achieved by other ensemble methods.
 
\textbf{Temporal Ensembles} train a single network and save that model's parameters at different points through time. The idea is that the model will converge to several diverse, yet accurate, local optima throughout its training cycle and the combination of those multiple optima will produce better results than the final model. Fast Committee Learning \cite{swann1998fast} first introduced this idea by simply choosing time slices through the training process. Recent methods seek better ways to choose which model checkpoints to save. Horizontal Voting Ensembles take the most recent states from a contiguous block of training epochs \cite{xie2013horizontal}, while Snapshot Ensembles use a cyclic learning rate schedule to alternate between converging to local minima and taking long jumps to new places in the parameter space \cite{huang2017snapshot}. Fast Geometric Ensembles improve on Snapshot Ensembles by looking for minima along high-accuracy pathways \cite{garipov2018loss}. FreeTickets saves states of a model trained with a sparse optimization algorithm \cite{liu2021freetickets}.

Temporal Ensembles yield good results with almost no extra training time compared to a single model. However, larger ensembles have diminishing returns. Model states taken early in the training process have poorer accuracy while model states taken later in the training process tend to be highly correlated.

\textbf{Evolutionary Ensembles} take a single network and generate explicit child networks according to a perturbative process. These methods have long been popular in the reinforcement learning domain as an alternative to gradient based optimization for small networks. Evolution Strategies (ES, CMA-ES, NES) generate populations of neural networks by adding random noise to the parameters of a parent \cite{salimans2017evolution, hansen2003reducing, wierstra2011natural}. Neuroevolution of Augmenting Topologies (NEAT) starts with a population of small networks and slowly complexifies them over time, incorporating concepts like speciation to encourage diversity \cite{stanley2002evolving}. MotherNets is a recent approach for deep neural networks that trains a small parent and hatches children via function preserving network morphisms \cite{wasay2020mothernets}. Child networks are created by adding additional layers and neurons on top of the core trained parent network.

Evolutionary Ensembles produce networks that are more diverse than Pseudo-Ensembles and Temporal Ensembles, at the cost of slower convergence and additional computation. Evolutionary Ensembles have historically been restricted to small network sizes due to the sample inefficiency of evolutionary optimization. However, they can generate new children cheaply, allowing for very large ensemble sizes. At this point, there has not been enough work applying evolutionary ensemble methods to deep vision networks to fully evaluate their potential.

\section{Prune and Tune Ensembles}

\textbf{The Parent Network:} Prune and Tune Ensembles don't require any special training considerations for the parent network. Any network architecture and optimization method can be used for the parent without modification. Common regularization techniques like batch normalization, weight decay and dropout are fully compatible with our method. We suggest following best practices for training the chosen network architecture. Furthermore, our approach can use pre-trained networks to leverage even greater cost savings in practical cases.

\textbf{Creating the Ensemble Members:} Child networks are created by making copies of the parent network and pruning the weights according to random or complementary ``anti-random" sampling methods. 
The topological transformations achieved with pruning allows each child network to explore an optimization space unique to its own architecture and they converge very fast due to their inherited parameters.

Modern pruning methods tend to work by deterministically selecting the best parameters to remove from a network. These approaches look at the magnitudes, importance coefficients or contributions to the gradient of the parameters \cite{blalock2020state, janowski1989pruning, lecun1990optimal, lee2019snip, frankle2019lottery}. While these methods produce very small and accurate single networks, we hypothesize that random pruning can be more effective when constructing diverse ensembles.

Due to the nature of extremely high dimensional neural networks, the probability is small that we will generate two child networks with a similar topology. In addition, our ensemble creation process is decoupled from the parent network's training process, which allows us to trivially create ensembles of any size.

\textit{Random Pruning:} These methods remove either connections, neurons, filters or layers randomly from neural networks. This is generally done in either a global or layer-wise fashion, which affects the distribution of pruned weights in networks that are unbalanced. In our experiments, we find little difference between global and layer-wise pruning. The optimal amount of pruning needed for the child networks is dependent on the parent network's size and architecture. We compare the impact of connection and neuron pruning on sparsity in our experimental section.

\textit{Anti-Random Pruning:}  A random pruning process can result in children that share portions of their respective parameter space. We introduce a method to improve upon random pruning that we call Anti-Random Pruning. Inspired by the concept of Anti-Random Testing \cite{malaiya1995antirandom}, we attempt to generate child networks that are maximally distant from one another while equally retaining all of the original parameters from the parent.

We perform this procedure using binary bit masks that represent pruned connections. We aim to maximize the total 
Cartesian distance between bit masks where the Cartesian distance for two binary vectors $A = \{a_1, ..., a_N\}$ and $B = \{b_1, ..., b_N\}$ is defined as:

\[
CD(A,B) = \sqrt{|a_1 - b_1| + ... + |a_N - b_N|}
\]

We wish to maximize the total distance $TD$ among all ensemble members $\{C_1, .., C_{N}\}$ such that:
\[ 
TD = \sum_{i=1}^{N} \sum_{j=1}^{N} CD(C_i, C_j)
\]

In practice, creating an ensemble that is maximally distant for all members would require an exhaustive search, which is infeasible for high dimensional neural networks. We suggest two alternative approaches that construct distant networks by using complementary masks and partitioning.

For any bit string $M$, the most distant counterpart is one in which the polarity of every bit is flipped: $M' =  1 - M$. We use this assumption to build an anti-random ensemble by first generating a random bit mask with 50\% sparsity. We construct the anti-network by taking the complement of the bit mask.  Each parameter of the parent model appears in only one of the two children.

The result is two child networks that are as distant from each other as possible while still inheriting parameters from the same parent. Given a parent network $W_p$ and a binary mask generated randomly with a 50\% sparsity target, $M = \{m_0, m_1, ..., m_n\}$, the resulting child networks are described as:

\[ C_1 = W_p \circ M ~~~~~~~~  \mbox{and} ~~~~~~~~ C_2 = W_p \circ M'. \]

This can be repeated $N$ times to create an ensemble of size $2N$. It is important to note that this approach requires 50\% sparsity to create ensembles members of equal size.

We extend this idea to larger parent networks using {\em anti-random partitioning}. For example, for an ensemble of size $N$, each child network can sample $\frac{1}{N}$ parameters from the parent network, without replacement.
This yields an ensemble $E$ that is a disjoint union (denoted $\coprod$) of its member networks $E = \coprod_{i} E_{i}.$  
Using this approach, every parameter in the original parent model is equally represented in the ensemble.

\textbf{Tuning The Ensemble Members:} Each child network inherits its parameters from an accurate parent network. 
Thus, child networks converge to good solutions after 
only a few training epochs. Due to the differences in the topology of each child network, the resulting models converge to unique places in the optimization space. 

\begin{figure}[t]
    \centering
    \includegraphics[width=\columnwidth]{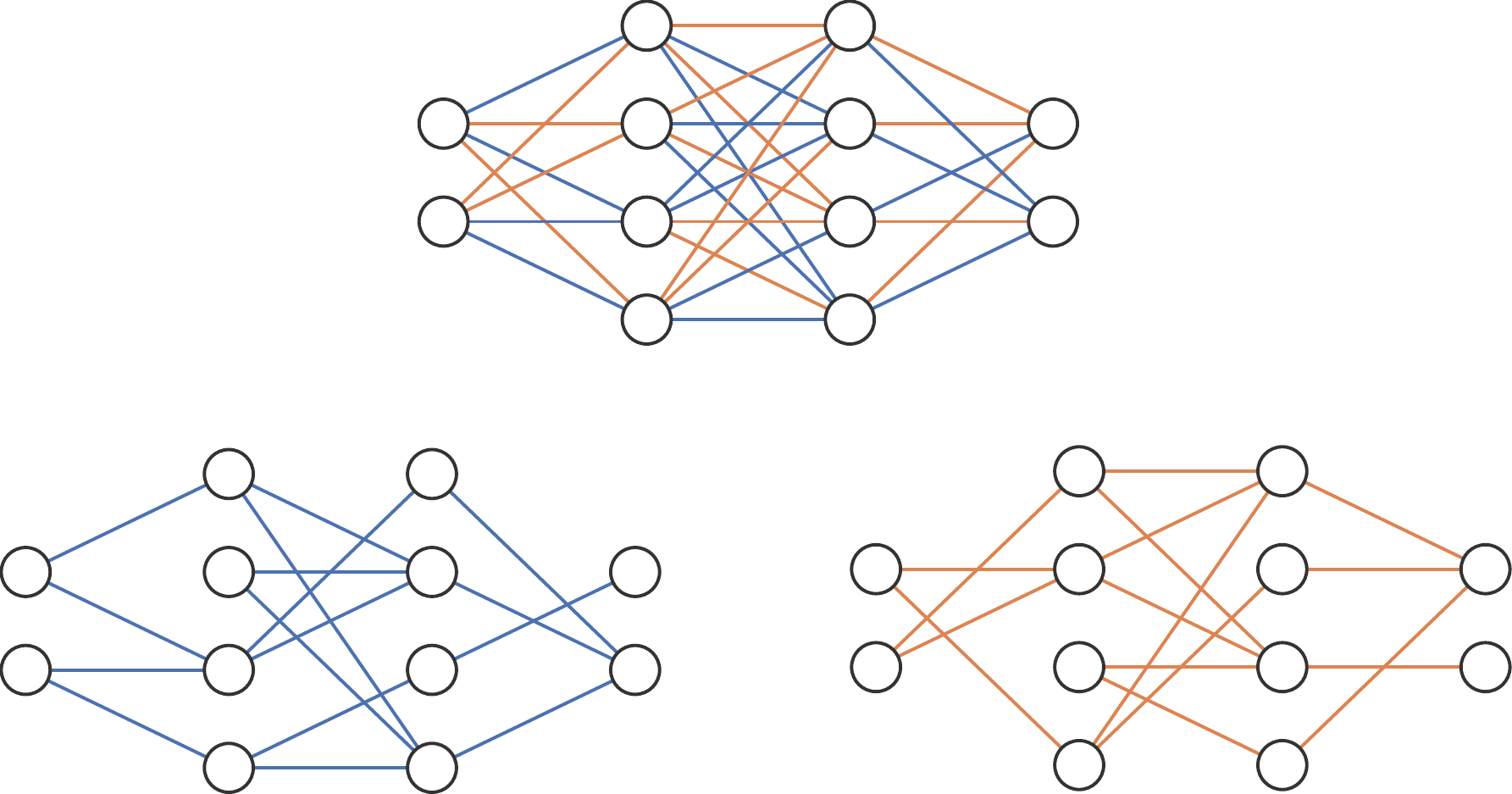}
    \caption{Parent network split into two child networks using Anti-Random Sampling. Every parameter in the parent's hidden layers appears in one of the two child networks exactly once.}
    \label{fig:antirandom-split}
\end{figure}

\textit{Bagging and Boosting}:
In order to encourage more diversity among ensemble members, it's possible to implement traditional ensemble techniques like bagging or boosting, where each child is tuned on a different subset of the training data.   In this paper we explore bagging \cite{breiman1996bagging},  where a child network is trained on 90\% 
of the available training data.  The subset composed of 90\% of the training data is randomly chosen for each child to promote diversity.

\textit{Traditional Fine Tuning}: This method for fine tuning the members of an ensemble involves additional training using a constant learning rate equal to the last learning rate used for the parent network. We use this method of tuning for small training budgets.

\textit{One-Cycle Tuning}: We use a one-cycle learning rate schedule \cite{smith2018superconvergence} to maximize diversity among child networks. This schedule is composed of three phases: a warm up, a cool down, and an annihilation. The warm up phase starts with a small learning rate that ramps up to a large learning rate. The cool down phase decays from the large learning rate down to a value several times smaller than the initial learning rate. The annihilation phase then decays the rate to 0. The one cycle policy encourages diversity among child networks as the large learning rates allow each child to move a greater distance from the parent network before converging.
The benefits of one-cycle tuning are more noticeable with larger training budgets \cite{le2021network}.

We use a revised one-cycle schedule that makes use of a two-phase cosine annealing instead of the original three-phase linear schedule \cite{le2021network}. We use a warm up phase for 10\% of the training budget which quickly ramps up from the lowest learning rate ($\eta_{min}$) used in the main training phase to the highest ($\eta_{max}$). The learning rate then decays to a very small value on the order of $1e-7$. The learning rate is updated per batch, where $T_{cur}$ is the current iteration and $T_{max}$ is the total number of iterations:

\[
\eta_t = \eta_{min} + \frac{1}{2}(\eta_{max} - \eta_{min})
(1 + cos(\frac{T_{cur}}{T_{max}} \pi))
\]

\textbf{Making Predictions:} The two most popular methods for combining model predictions in ensembles include majority vote and weighted model averaging \cite{fragoso2017bayesian, dzeroski2004combining, vanderlaan2007super}. We use model averaging, which is standard practice for comparable ensemble methods on image classification \cite{huang2017snapshot, garipov2018loss, liu2021freetickets, havasi2021training}.

To achieve better normalization, we average the softmax of each ensemble member's outputs to ensure that all ensemble members produce outputs of the same scale.
\[ y_e = argmax( \frac{1}{S} \sum_{i=1}^{S} \sigma(f_i(x))) \]
where $y_e$ is the ensemble prediction, $S$ is the number of members in the ensemble (the ensemble size), $\sigma$ is the softmax function and $f_i(x)$ is the output of the individual ensemble member $i$.

\section{Experimental Configurations}

\begin{figure*}[t]
    \centering
    \includegraphics[width=.52\columnwidth]{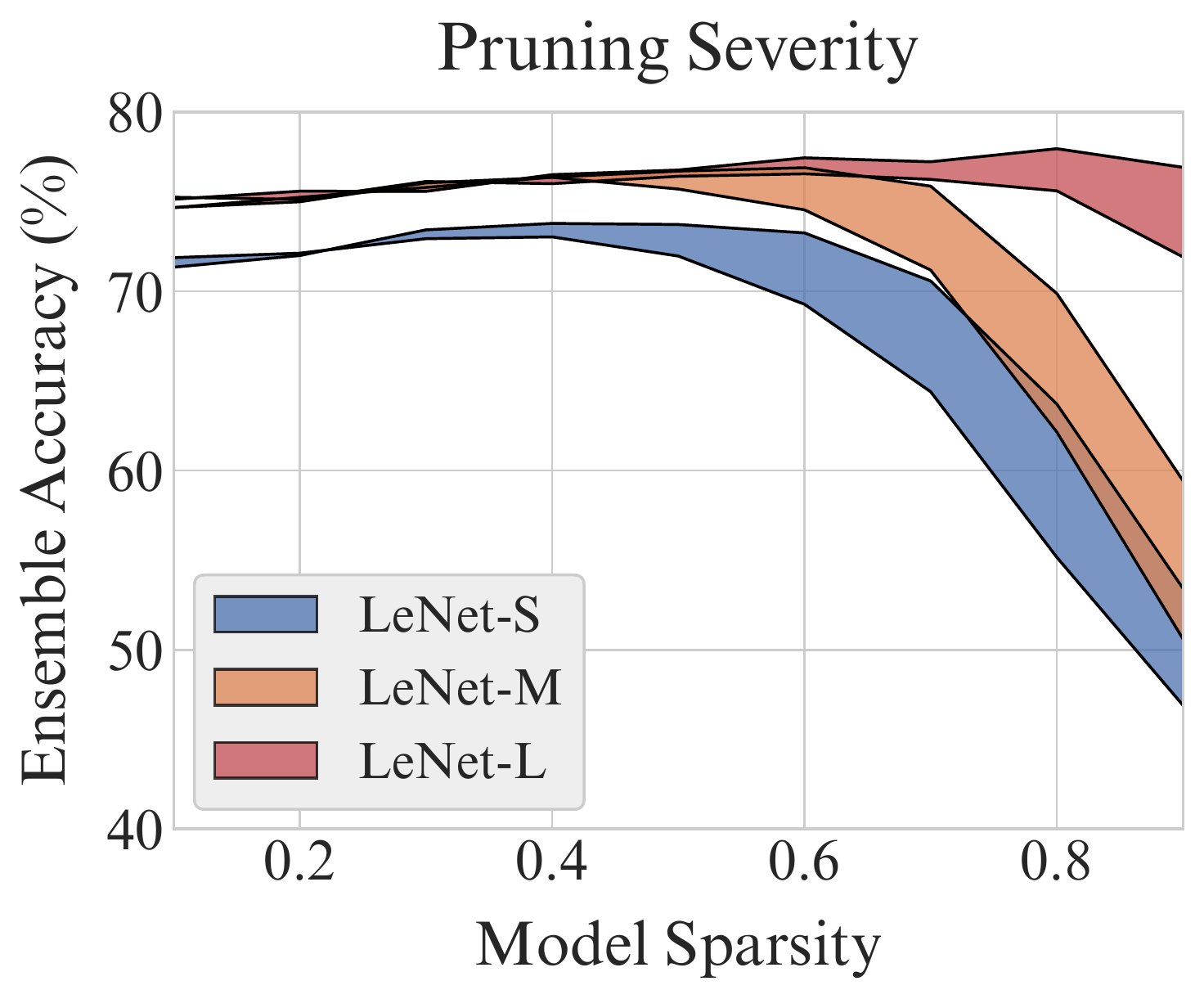}
    \includegraphics[width=.52\columnwidth]{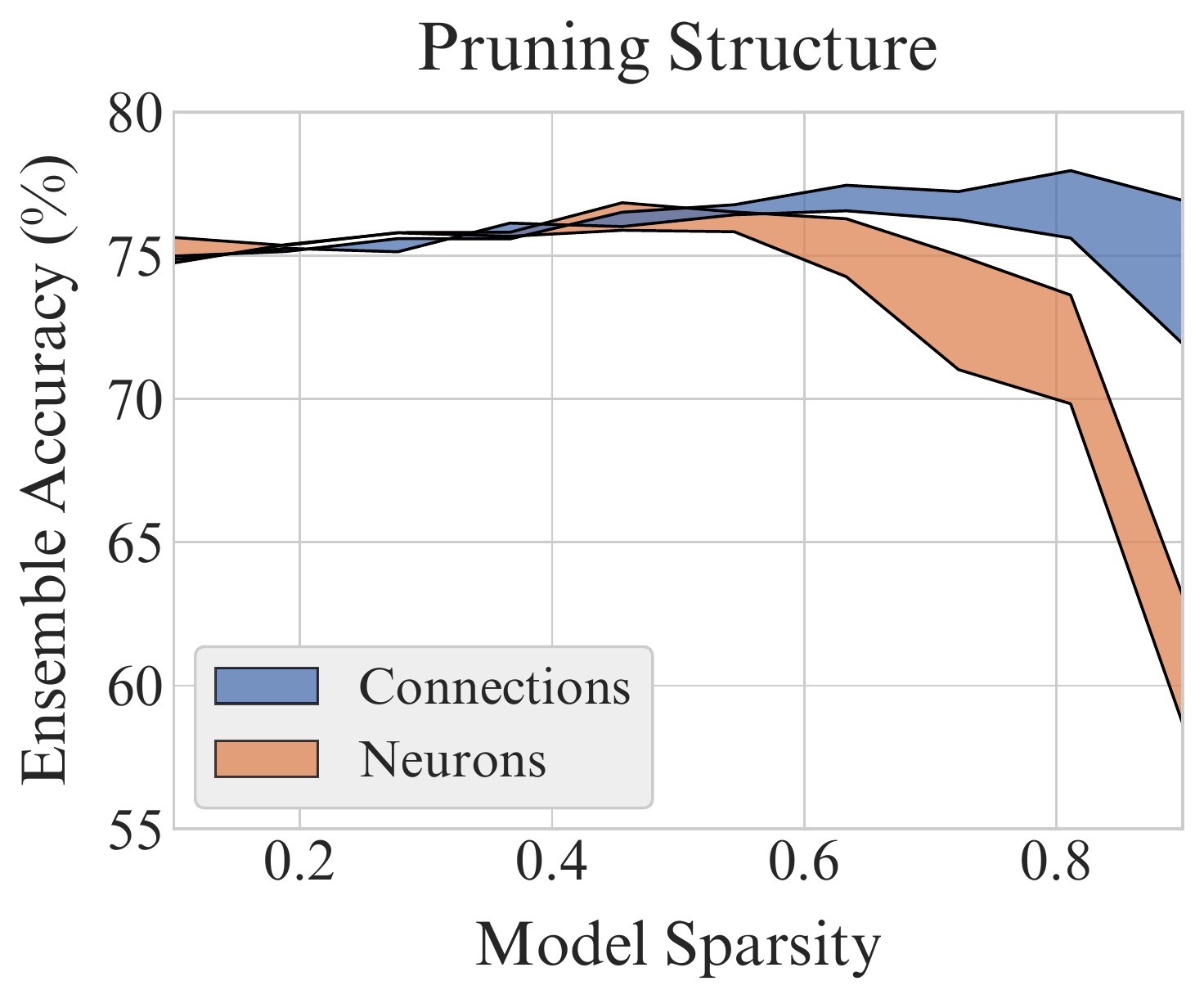}
    \includegraphics[width=.52\columnwidth]{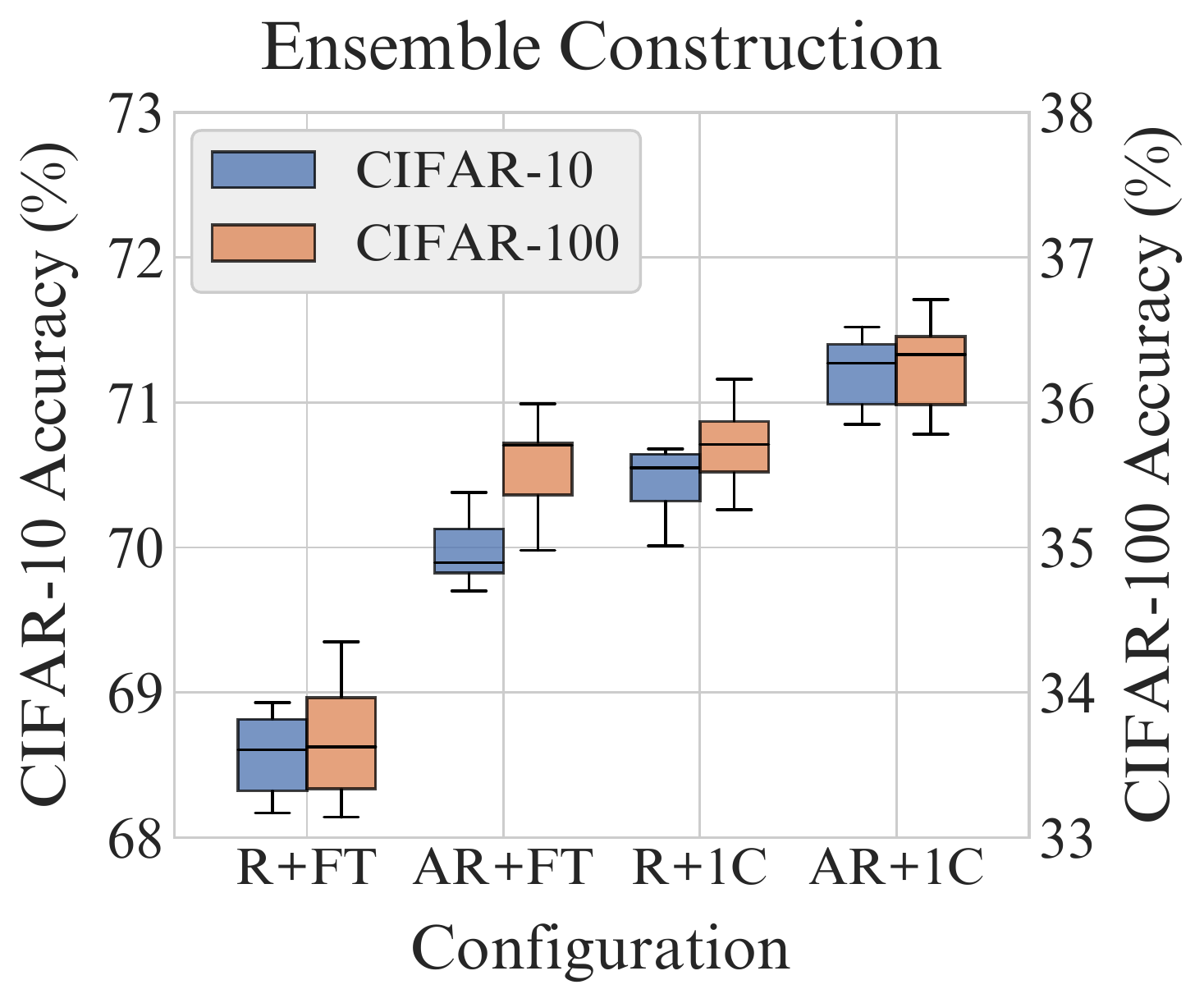}
    \includegraphics[width=.52\columnwidth]{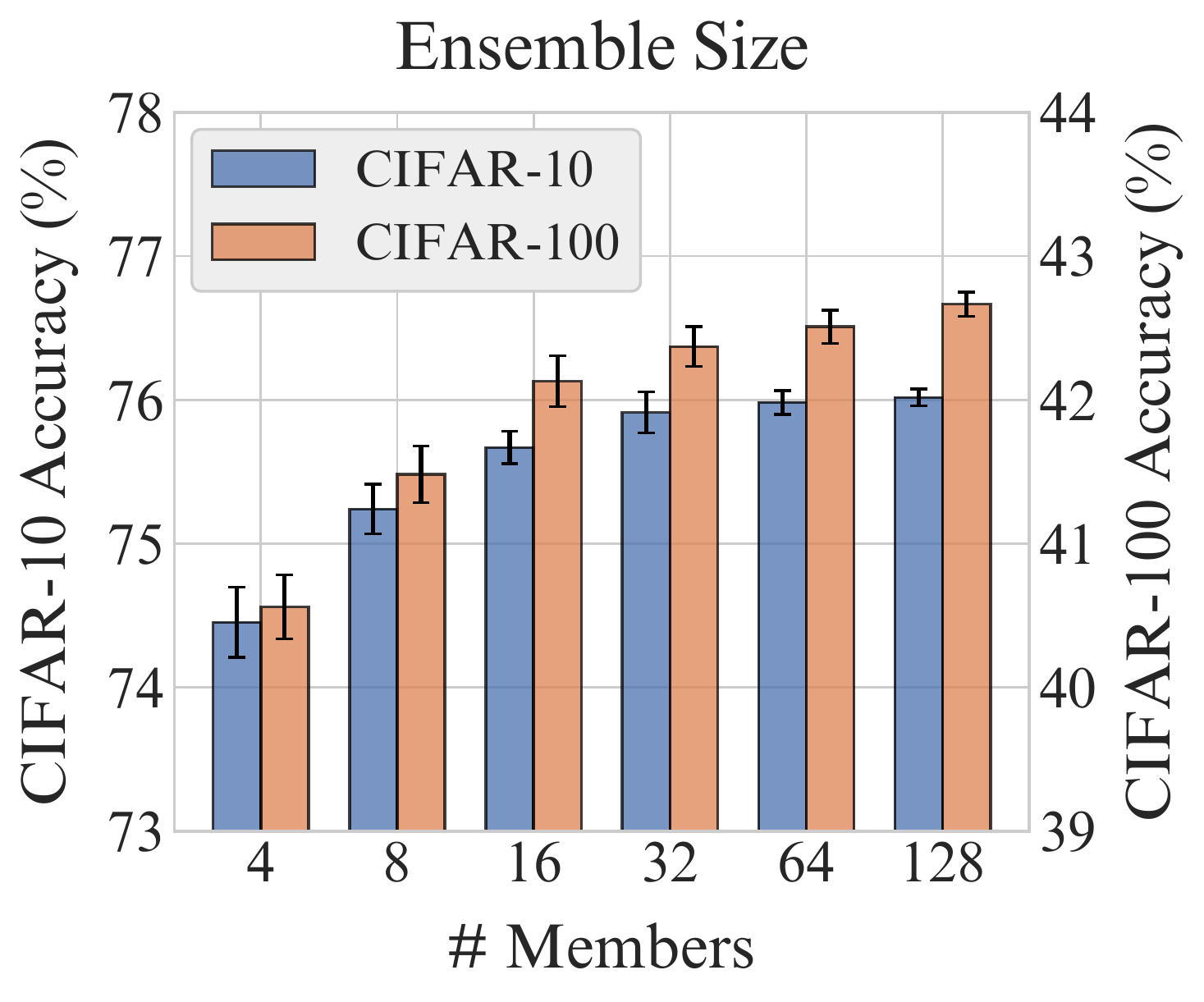}

\caption{Hyperparameter ablations on CIFAR-10/CIFAR-100 with various LeNet configurations. The first figure reports the effect of sparsity on ensemble performance for models of different size. The second compares neuron and connection elimination for the large model. The third figure explores both random/anti-random pruning and constant/cyclic tuning schedules over 10 runs. The final figure evaluates the effect of ensemble size up to $M=128$.}

    \label{fig:ablations}
\end{figure*}

All experiments are designed to be accessible and easily replicable with affordable hardware. Datasets and models are open source and linked in the appendix. 

\textbf{Datasets}: We use the computer vision datasets, CIFAR-10 and CIFAR-100 \cite{krizhevsky2012learning}. CIFAR consists of 60,000 small natural colored images of 32x32 pixels in size. Those 60,000 images are split up into 50,000 training images and 10,000 testing images. CIFAR-10 samples from 10 classes of images, while CIFAR-100 samples from 100 classes of images. CIFAR-100 is more difficult than CIFAR-10 as each class will have only 500 training samples compared to 5,000 in CIFAR-10. In addition, we evaluate our large experiments on corrupted versions of CIFAR-10 and CIFAR-100 \cite{hendrycks2019robustness}. These additional test sets are generated by adding 20 different kinds of image corruptions (gaussian noise, snow, blur, pixelation, etc.) at five different levels of severity to the original CIFAR test sets. The total number of images in each of these additional sets is 1,000,000.

\textbf{Models}: We use a variety of popular deep convolutional architectures, chosen both for historical significance as well as popularity and performance.

\textit{LeNet} \cite{lecun1989backpropagation}: LeNet is one of the earliest successful convolutional architectures. LeNet-5 consists of two convolutional layers, two fully connected dense layers, and one output layer. 
We make several adjustments that are common practice in modern vision networks. These changes include the use of ReLU for the activation functions and the use of maxpooling layers after the convolutional layers. LeNet-5 contains a total of $\sim$ 140,000 parameters.

\textit{ResNet} \cite{he2015deep}: ResNets are ubiquitious and have become one of the most popular vision architectures in deep learning. ResNets can produce significantly deeper neural networks by utilizing residual blocks and shortcut connections between groups of layers. We use ResNet-18, an 18 layer version introduced in the original ResNet paper. This variant contains $\sim$ 11 million parameters.

\textit{DenseNet} \cite{huang2018densely}: DenseNets build upon the ideas of shortcut connections introduced in the ResNet paper. DenseNet directly connects each layer to every other successive layer within a dense block. At each layer, the inputs are a concatenation of all previous layer outputs. We use DenseNet-121, a variant that contains 121 layers in total and is the smallest architecture introduced in the original DenseNet paper. This variant contains  $\sim$ 7 million parameters.

\textit{Wide ResNet} \cite{zagoruyko2017wide}: Wide ResNets were introduced as a variant to the original ResNet, addressing the problem of diminishing feature reuse in deep residual networks. Wide ResNets increase the width of residual blocks and show excellent performance with significantly fewer layers than their ResNet counterparts. We use WRN-28-10, a 28 layer variant introduced in the original Wide ResNet paper. This model contains $\sim$ 36 million parameters.

\textbf{Hardware}: All models are trained on a single Nvidia GTX-1080-Ti GPU.

\section{Experimental Results}

The following sections detail several experiments conducted in order to evaluate the efficacy of Prune and Tune Ensembles at very different scales. 

First, we perform several hyperparameter ablations and evaluate the impact of those choices on generalization using variations of the LeNet model. 

Second, we compare Prune and Tune Ensembles to alternative methods on CIFAR-10 and CIFAR-100 using an extremely small training budget of only 16 total epochs.

Third, we compare Prune and Tune Ensembles to state-of-the-art low cost ensemble learning methods with a large training budget of 200 total epochs. Along with accuracy, we report uncertainty estimations on corrupted versions of the CIFAR datasets \cite{nado2021uncertainty, hendrycks2019robustness}.

\subsection{Hyperparameter Ablations}

We explore the impact of several hyperaparameters on our ensembles, including: amount of sparsity, random/anti-random pruning, connection/neuron elimination, constant/cyclic tuning schedule and ensemble size. We use variations of the LeNet-5 model architecture where we vary the number of neurons in each hidden layer. We call these small, medium and large models the {\it LeNet-S} (253,290 parameters), {\it LeNet-M} (1,007,306 parameters), and {\it LeNet-L} versions (4,017,546 parameters). See appendix for additional details on the layer configurations.

We conduct our experiments on CIFAR-10 and CIFAR-100. We use Stochastic Gradient Descent (SGD) with Nesterov momentum for our evaluation of random/anti-random pruning and constant/cyclic tuning schedule. We use an initial learning rate of $\eta_1 = 0.1$ for 50\% of the training budget which decays linearly to $\eta_2 = 0.001$ at 90\% of the training budget. The learning rate is kept constant at $\eta_2 = 0.001$ for the final 10\% of training. Children are tuned with either a constant learning rate of $\eta=0.01$ for 5 epochs or with a one-cycle schedule that ramps up from $\eta_1=0.001$ to $\eta_2=0.1$ at 10\% of tuning, then decaying to $\eta_3=1e-7$ at the end of 5 epochs. For all other experiments, we use ADAM with a learning rate of $\eta = 0.001$.

To evaluate sparsity, we take each model variation ({\it LeNet-S}, {\it LeNet-M}, and {\it LeNet-L}) and train the parent network for 8 epochs. An ensemble of 8 children are created that are each tuned for 3 epochs. We report the ensemble accuracy at both 1 and 3 epochs of tuning for child sparsity ranging from 10\% to 90\%. 
All models are pruned using global connection elimination. 

To evaluate neuron and connection elimination, we train the {\it LeNet-L} parent network for 8 epochs. An ensemble of 8 children is created, where each is pruned using either connection or neuron elimination. Each child is tuned for 3 epochs. We report ensemble accuracy for both neuron and connection elimination at various levels of sparsity between 10\% and 90\%.

To evaluate ensemble size, we take the {\it LeNet-M} network and produce up to 128 candidate member networks using 50\% sparsity connection pruning and 1 epoch of constant rate tuning ($\eta=0.001$). We evaluate ensemble performance over 10 runs as we increase the size of the ensemble from 4 to 128.

We finally evaluate random/anti-random pruning and constant/cyclic tuning. We use the {\it LeNet-L} network and create an ensemble of two models using each combination of random (R) or anti-random (AR) pruning and constant fine-tuning (FT) or one-cycle tuning (1C). We record results for 10 sample runs of each configuration.

Figure 3 shows the results of all ablation experiments. The leftmost figure illustrates the capacity that larger models have for learning more efficiently. Ensemble accuracy improves up to a critical region as child networks become more sparse, {\em without any additional epochs of training}. This indicates that more sparsity can yield greater diversity in the ensemble. 

The second figure shows that connection pruning allows for more sparsity when compared to neuron pruning. We see no difference in performance for sparsity values up to 50\%, but we note that the performance collapses for neuron pruning much sooner than for connection pruning.

The third figure shows that the combination of anti-random pruning and one-cycle tuning outperforms other configurations. We see significant evidence of enhanced diversity between two models with the addition of these two techniques on both CIFAR-10 and CIFAR-100.

The rightmost figure shows that ensembles consistently perform better as we increase the number of ensemble members. These results suggest significant advantages in using methods that can dynamically generate ensemble members, as performance monotonically increases with size.

\subsection{Comparisons for Small Training Budgets}

Ensemble learning literature has not focused on evaluating approaches for very small training budgets, but we believe there is significant value in doing so. We take inspiration from Stanford's DAWNBench experiment, the goal of which is to train a single model to a target accuracy in as little time as possible \cite{coleman2017DAWNBench}. Thus, we conduct a comparison among methods that limits the total training budget to only 16 epochs.
Each method is ran with a sample size of 10 runs, except for both versions of Prune and Tune, which are ran 30 times each.

The {\bf Independent Model} is a standard neural network model trained for all 16 epochs.
The {\bf Bagged Ensemble} contains 8 full size models, each trained for two epochs. Each member is trained on a 90\% sample of the full training set.
The {\bf Dropout Model} is identical to the Independent Model except dropout layers are inserted after every convolutional and linear layer.
These dropout layers use the same 50\% sparsity value that we use for Prune and Tune, to keep the number of parameters consistent. For {\bf Prune and Tune}, we train a single parent network for 8 epochs, we then create 8 children using random connection pruning with a 50\% sparsity target. For {\bf Prune and Tune (AR)} we create 8 children using anti-random (AR) pruning: we create 4 pairs of network/anti-networks each with 50\% sparsity.  We fine tune each child for only 1 epoch. Each approach uses the ADAM optimizer with a fixed learning rate of $\eta=0.001$.

We report the mean accuracy and standard error achieved for each method in Table 1. One may be surprised by the poor performance of Dropout. A 50\% dropout rate is larger than typical configurations, but we chose to do this to closely match the configuration of Prune and Tune Ensemble children. This combined with the small budget leads to a much slower convergence for these models. This experiment is designed to create an initial baseline for small training budgets that can be further explored in future work.

In these small fixed training budget experiments, comparisons using a simple t-test shows that the Prune and Tune ensemble methods are significantly better than all other methods (at p=0.0001) over 30 runs.

\begin{table}[t]
\small

\begin{tabularx}{\columnwidth}{l X c c}
\toprule
Model & Method & CIFAR-10 & CIFAR-100 \\
\midrule
LeNet-L  & Independent & 70.29 $\pm$ 0.21 & 34.39 $\pm$ 0.18 \\
        & Dropout & 70.06 $\pm$ 0.32  & 28.98 $\pm$ 0.33 \\
        & Bagged & 70.97 $\pm$ 0.13  & 32.40 $\pm$ 0.11 \\
        & PAT (R) & 75.51 $\pm$ 0.06  & 41.68 $\pm$ 0.14 \\
        & PAT (AR) & \textbf{75.75} $\pm$ \textbf{0.06} & \textbf{41.85} $\pm$ \textbf{0.13} \\
\midrule
ResNet-18 & Independent & 80.99 $\pm$ 0.11 & 49.26 $\pm$ 0.09 \\
        & Dropout & 72.97 $\pm$ 0.41 & 37.69 $\pm$ 0.28 \\
        & Bagged & 79.88 $\pm$ 0.05 & 46.18 $\pm$ 0.06 \\
        & PAT (R) & 84.61 $\pm$ 0.07  & 57.08 $\pm$ 0.07 \\
        & PAT (AR) & \textbf{84.68} $\pm$ \textbf{0.04} & \textbf{57.26} $\pm$ \textbf{0.06} \\
\midrule
DenseNet-121 & Independent & 84.14 $\pm$ 0.09 & 56.62 $\pm$ 0.09 \\
          & Dropout & 79.59 $\pm$ 0.36 & 44.02 $\pm$ 0.58 \\
          & Bagged & 81.16 $\pm$ 0.04 & 48.48 $\pm$ 0.06 \\
          & PAT (R) & 86.26 $\pm$ 0.07  & 62.02 $\pm$ 0.07 \\
            & PAT (AR) & \textbf{86.35} $\pm$ \textbf{0.04} & \textbf{62.09} $\pm$ \textbf{0.06} \\
\bottomrule
\end{tabularx}

\caption{Accuracy of various methods for 16 epochs of training. Ensembles contain eight models.}\label{tab:low-budget-table}
\end{table}

\subsection{Comparisons for Large Training Budgets}

\begin{table*}[t]
\begin{tabularx}{\textwidth}{X l l l l l l c c}
\toprule
Methods (CIFAR-10/WRN-28-10) & Acc $\uparrow$ & NLL $\downarrow$ & ECE $\downarrow$ & cAcc $\uparrow$ & cNLL $\downarrow$ & cECE $\downarrow$ & FLOPs $\downarrow$ & Epochs $\downarrow$ \\
\midrule
Independent Model$^*$ & 96.0 & 0.159 & 0.023 & 76.1 & 1.050 & 0.153 & 3.6e17 & 200 \\
Monte Carlo Dropout$^*$ & 95.9 & 0.160 & 0.024 & 68.8 & 1.270 & 0.166 & 1.00x & 200 \\
TreeNet (M=3)$^*$ & 95.9 & 0.258 & 0.018 & 75.5 & {\bf 0.969} & 0.137 & 1.52x & 250 \\
SSE (M=5) & 96.3 & 0.131 & 0.015 & 76.0 & 1.060 & 0.121 & 1.00x & 200 \\
FGE (M=12) & 96.3 & 0.126 & 0.015 & 75.4 & 1.157 & 0.122 & 1.00x & 200 \\
PAT (M=6) (AR + 1C) & {\bf 96.48} & {\bf 0.113} & {\bf 0.005} & {\bf 76.23} & 0.972 & {\bf 0.081} & {\bf 0.85x} & {\bf 200} \\
\midrule
BatchEnsemble (M=4)$^*$ & 96.2 & 0.143 & 0.021 & 77.5 & 1.020 & 0.129 & 4.40x & 250 \\
MIMO (M=3)$^*$ & 96.4 & 0.123 & 0.010 & 76.6 & 0.927 & 0.112 & 4.00x & 250 \\
EDST (M=7)$^*$ & 96.4 & 0.127 & 0.012 & 76.7 & 0.880 & 0.100 & 0.57x & 850 \\
DST (M=3)$^*$ & 96.4 & 0.124 & 0.011 & 77.6 & 0.840 & 0.090 & 1.01x & 750 \\
Dense Ensemble (M=4)$^*$ & 96.6 & 0.114 & 0.010 & 77.9 & 0.810 & 0.087 & 1.00x & 800 \\
\bottomrule
\end{tabularx}

\vspace{0.2in}

\begin{tabularx}{\textwidth}{X l l l l l l c c }
\toprule
Methods (CIFAR-100/WRN-28-10) & Acc $\uparrow$ & NLL $\downarrow$ & ECE $\downarrow$ & cAcc $\uparrow$ & cNLL $\downarrow$ & cECE $\downarrow$ & FLOPs $\downarrow$ & Epochs $\downarrow$ \\
\midrule
Independent Model$^*$ & 79.8 & 0.875 & 0.086 & 51.4 & 2.700 & 0.239 & 3.6e17 & 200 \\
Monte Carlo Dropout$^*$ & 79.6 & 0.830 & 0.050 & 42.6 & 2.900 & 0.202 & 1.00x & 200 \\
TreeNet (M=3)$^*$ & 80.8 & 0.777 & 0.047 & {\bf 53.5} & {\bf 2.295} & 0.176 & 1.52x & 250 \\
SSE (M=5) & 82.1 & 0.661 & 0.040 & 52.2 & 2.595 & 0.145 & 1.00x & 200 \\
FGE (M=12) & 82.3 & 0.653 & 0.038 & 51.7 & 2.638 & 0.137 & 1.00x & 200 \\
PAT (M=6) (AR + 1C) & {\bf 82.67} & {\bf 0.634} & {\bf 0.013} & 52.70 & 2.487 & {\bf 0.131} & {\bf 0.85x} & {\bf 200} \\
\midrule
BatchEnsemble (M=4)$^*$ & 81.5 & 0.740 & 0.056 & 54.1 & 2.490 & 0.191 & 4.40x & 250 \\
MIMO (M=3)$^*$ & 82.0 & 0.690 & 0.022 & 53.7 & 2.284 & 0.129 & 4.00x & 250 \\
EDST (M=7)$^*$ & 82.6 & 0.653 & 0.036 & 52.7 & 2.410 & 0.170 & 0.57x & 850 \\
DST (M=3)$^*$ & 82.8 & 0.633 & 0.026 & 54.3 & 2.280 & 0.140 & 1.01x & 750 \\
Dense Ensemble (M=4)$^*$ & 82.7 & 0.666 & 0.021 & 54.1 & 2.270 & 0.138 & 1.00x & 800 \\
\bottomrule
\end{tabularx}

\caption{Results for ensembles of WideResNet-28-10 models on both CIFAR-10 and CIFAR-100. Methods with $^*$ denote results obtained from \cite{havasi2021training, liu2021freetickets}. Best low-cost ensemble results are {\bf bold}. cAcc, cNLL, and cECE correspond to corrupted test sets. We report the mean values over 10 runs for PAT.}
\end{table*}

We next evaluate Prune and Tune Ensembles with a larger training budget of 200 epochs with WideResNet-28-10. We take the training configuration, ensemble size and parameter setting directly from studies of three state-of-the-art low-cost ensemble methods: MotherNets \cite{wasay2020mothernets}, Snapshot Ensembles \cite{huang2017snapshot}, and Fast Geometric Ensembles \cite{garipov2018loss}. We also compare our results with published results of several recent low-cost ensemble methods including: TreeNets \cite{lee2015m}, BatchEnsemble \cite{wen2020batchensemble}, FreeTickets \cite{liu2021freetickets}, and MIMO \cite{havasi2021training}.

All methods compared use WideResNet-28-10 and Stochastic Gradient Descent with Nesterov momentum $\mu=0.9$ and weight decay $\gamma=0.0005$ \cite{sutskever13nesterov}. The Prune and Tune ensemble size and training schedule is as used in previous comparisons \cite{wasay2020mothernets, garipov2018loss}.  We use a batch size of 128 for training and use random crop, random horizontal flip, and mean standard scaling data augmentations for all approaches \cite{garipov2018loss, havasi2021training, liu2021freetickets, huang2017snapshot}. The parent learning rate uses a step-wise decay schedule. An initial learning rate of $\eta_1=0.1$ is used for 50\% of the training budget which decays linearly to $\eta_2=0.001$ at 90\% of the training budget. The learning rate is kept constant at $\eta_2=0.001$ for the final 10\% of training. 

The {\bf Independent Model} is a single WideResNet-28-10 model trained for 200 Epochs. The {\bf Dropout Model} includes dropout layers between convolutional layers in the residual blocks at a rate of 30\% \cite{zagoruyko2017wide}. {\bf Snapshot Ensembles} (SSE) use a cosine annealing learning rate with an initial learning rate $\eta = 0.1$ for a cycle length of 40 epochs \cite{huang2017snapshot}. {\bf Fast Geometric Ensembles} (FGE) use a pre-training routine for 156 epochs. A curve finding algorithm then runs for 22 epochs with a cycle length of 4, each starting from checkpoints at epoch 120 and 156. {\bf TreeNets}, \cite{lee2015m}, {\bf BatchEnsemble} \cite{wen2020batchensemble} and {\bf MIMO} \cite{havasi2021training} are all trained for 250 epochs. {\bf FreeTickets} introduces several configurations for building ensembles. We include their two best configurations for Dynamic Sparse Training (DST, M=3, S=0.8) and Efficient Dynamic Sparse Training (EDST, M=7, S=0.9).

{\bf Prune and Tune Ensembles} (PAT) train a single parent network for 140 epochs. Six children are created with anti-random pruning (50\% sparsity) and tuned with a one-cycle learning rate for 10 epochs. The tuning schedule starts at $\eta_1=0.001$, increases to $\eta_2=0.1$ at 1 epoch and then decays to $\eta_3=1e-7$ using cosine annealing for the final 9 epochs.

There is the potential to further improve performance by exploring larger ensemble sizes and different ratios of parent to child training time.

We report the mean accuracy (Acc), negative log likelihood (NLL), and expected calibration error (ECE) over 10 runs on both CIFAR-10 and CIFAR-100 along with their corrupted variants \cite{nado2021uncertainty, hendrycks2019robustness}. We also report the total number of floating point operations (FLOPs) and epochs used for training each method. We organize our tables into two groups based on training cost. The first group consists of low-cost training methods that take approximately as long as a single network would take to train. The second group of methods use either significantly more epochs or compute per epoch to achieve comparable performance. MIMO and BatchEnsemble both make use of batch repetition to train on more data while keeping the number of epochs low. FreeTickets (DST and EDST) use very sparse networks to keep FLOP counts low while using many more training epochs.

Prune and Tune Ensembles (using anti-random pruning and one-cycle tuning) outperform all low-cost ensemble methods and are competitive with methods that train for significantly longer. Our approach produces well calibrated, robust and diverse ensembles with excellent performance on out of distribution corrupted datasets.

\section{Conclusions}

The Prune and Tune Ensemble algorithm is flexible and enables the creation of diverse and accurate ensembles at a significantly reduced computational cost. We do this by 1) training a single large parent network, 2) creating child networks by copying the parent and significantly pruning them using random or anti-random sampling strategies, and 3) fine tuning each of the child networks for a small number of training epochs.

With the experiments introduced here, Prune and Tune Ensembles outperform low-cost ensembling approaches on benchmark image classification datasets with a variety of popular deep neural network architectures. Our approach achieves comparable accuracy to ensembles that are trained on 4-5x more data. Prune and Tune Ensembles not only improve upon the training-time/accuracy trade-off, but also reduce memory and computational cost thanks to the use of sparse child networks.

Prune and Tune Ensembles can offer a new lens through which we can analyze generalization and diversity among subcomponents of deep neural networks. We hope to further explore larger scale benchmarks, anti-random pruning methodologies and applications to new problem domains in future work.

\section{Acknowledgments}

This work is supported by National Science Foundation award IIS-1908866.

\bibliography{bibliography}

\appendix

\section{Appendix}

\subsection{Pseudocode}

$W_p$ represents the parent network's weights, $W_{Ci}$ represents a child network's weights, $X$ is the full training data, $\hat{X}$ is the training data given to ensemble members, $M_i$ is a bit mask to represent which parameters to prune, and $E$ is a container to hold the resulting ensemble members.

\begin{algorithm}[!h]
\SetAlgoLined
 \KwIn{N, the number of ensemble members}
 $W_P \leftarrow initialize()$ \\ 
 $W_P \leftarrow trainToConvergence(f(X; W_P))$ \\ 
 $E \leftarrow []$ \\ 
 
 \For{i in 1 to N}{
    $M_i \leftarrow (0+1)^{|W_P|}$ \\
    $W_{Ci} \leftarrow prune(M_i, W_P)$ \\
    $W_{Ci} \leftarrow tune(f(\hat{X}; W_{Ci}))$ \\
    $E \leftarrow insert(E, W_{Ci})$ \\
 }
 \KwOut{E}
 \caption{Prune and Tune Ensemble Algorithm}
\end{algorithm}

\subsection{Codebase}

Our codebase is open source and MIT licensed. The repository for reproducing our experiments can be found at \url{https://github.com/tjwhitaker/prune-and-tune-ensembles}.

\subsection{Open Source Implementations}

We use PyTorch and the following open source implementations for all of our model implementations.

\begin{itemize}
\item ResNet-18 (MIT Licensed) \url{https://github.com/huyvnphan/PyTorch_CIFAR10/blob/master/cifar10_models/resnet.py}

\item DenseNet-121 (MIT Licensed) \url{https://github.com/huyvnphan/PyTorch_CIFAR10/blob/master/cifar10_models/densenet.py}

\item WideResNet-28-10 (MIT Licensed) \url{https://github.com/meliketoy/wide-resnet.pytorch/blob/master/networks/wide_resnet.py}

\item We also use the following repository for investigating the loss landscapes of pruned children. (MIT Licensed) \url{https://github.com/tomgoldstein/loss-landscape}
\end{itemize}

\subsection{LeNet-5 Model Configurations}

We use three variations of the LeNet model in our ablation experiments. Table 3 details the number of units in each of the hidden layers. The number of units is doubled for each variation of the network, and the total number of parameters roughly increases by 4x. We make several adjustments that are common practice in modern vision networks. These include using ReLU for the activation functions instead of tanh and the  addition  of  max pooling after the convolutional layers instead of average pooling.

\begin{table}[t]
\centering
\begin{tabularx}{\columnwidth}{ X c c c } 
\toprule
 ~ & & Model Variations & \\
\cmidrule{2-4}
Layer & LeNet-S & LeNet-M & LeNet-L \\
\midrule
Conv1 &  16 & 32 & 64 \\ 
Conv2 &  32 & 64 & 128 \\
FC1 & 256 & 512 & 1024 \\
FC2 & 128 & 256 & 512 \\
\midrule
\# Params & 253,290 & 1,007,306 & 4,017,546 \\
\bottomrule
\end{tabularx}
\caption{LeNet-5 model configurations. Values represent the number of {\it units} or {\it neurons} in each layer for each model variation.}
\end{table}

\subsection{Additional Results}

{\bf Standard Deviations:} We could not find standard deviations or mentions of sample sizes for approaches reported in our large budget experiment. Due to space constraints, we include ours here in Table 4 as supplementary information. We report our results collected over 10 runs. C-CIFAR-10 and C-CIFAR-100 are the corrupted test sets of CIFAR-10 and CIFAR-100 \cite{hendrycks2019robustness}.

\begin{table}
\small
\begin{tabularx}{\columnwidth}{X c c c}
\toprule
Dataset & Acc. $\uparrow$ & NLL $\downarrow$ & ECE $\downarrow$ \\
\midrule
CIFAR-10 & $96.48^{\pm 0.06}$ & $0.113^{\pm 0.002}$ & $0.005^{\pm 0.001}$  \\
C-CIFAR-10 & $76.23^{\pm 0.19}$ & $0.969^{\pm 0.022}$ & $0.081^{\pm 0.004}$  \\
CIFAR-100 & $82.67^{\pm 0.07}$ & $0.634^{\pm 0.003}$ & $0.013^{\pm 0.002}$  \\
C-CIFAR-100 & $52.70^{\pm 0.16}$ & $2.487^{\pm 0.047}$ & $0.131^{\pm 0.004}$  \\
\bottomrule
\end{tabularx}
\caption{Mean $\pm$ Standard Deviation for Prune and Tune Ensembles (AR + 1C) (M=6) over 10 runs. Experimental configuration is described in the large budget experiment.}
\end{table}

{\bf Late Phase Weights:} This is a recent method that appears to be very competitive with current state-of-the-art algorithms \cite{vonoswald2021neural}. They report test accuracy and negative log likelihoods for CIFAR-10, CIFAR-100 and corrupted CIFAR-100. They do not report estimated calibration error and uncertainty estimations on CIFAR-10, which is the primary reason the comparison did not make it into our large budget experiment in the main paper.

Table 5 details the results for WideResNet-28-10. Both approaches train for 200 epochs using stochastic gradient descent with Nesterov momentum $\mu=0.9$ and weight decay $\gamma=0.0005$ \cite{sutskever13nesterov}. We use a batch size of 128 for training and use random crop, random horizontal flip, and mean standard scaling data augmentations. The parent learning rate uses a step-wise decay schedule. An initial learning rate of $\eta_1=0.1$ is used for 50\% of the training budget which decays linearly to $\eta_2=0.001$ at 90\% of the training budget. The learning rate is kept constant at $\eta_2=0.001$ for the final 10\% of training.

\begin{table}
\centering
\begin{tabularx}{\columnwidth}{ X c c c } 
\toprule
 Method & CIFAR-10 & CIFAR-100 & C-CIFAR-100 \\
\midrule
LPBN$^*$ & $96.46^{\pm 0.15}$ & $82.87^{\pm 0.22}$ & $47.84^{\pm 0.41}$ \\
PAT & $96.48^{\pm 0.06}$ & $82.67^{\pm 0.07}$ & $52.70^{\pm 0.16}$ \\
\bottomrule
\end{tabularx}
\caption{Comparison of Late Phase Weights (M=10) and Prune and Tune Ensembles (M=6) with WideResNet-28-10 on CIFAR-10, CIFAR-100 and Corrupted CIFAR-100. $^*$ denotes results for Late Phase Batch Norm (LPBN) \cite{vonoswald2021neural}.}
\end{table}

\subsection{Diversity of Model Predictions}


Ensemble performance increases with the number of well trained and diverse models it contains \cite{bonab2016theoretical, oshiro2012forest}. We explore and compare the diversity of our ensembles with several other ensemble learning methods and with several different measures that compare the pairwise outputs of ensemble members, including correlation, disagreement ratio and Kullback-Leibler divergence.

\subsubsection{Pairwise Output Correlation}

We start with a measure of the correlation between outputs of any two models in the ensemble. We perform this for each sample in the test set and average their scores.
\[
    d_{corr}(f_1,f_2) = \frac{1}{N} \sum_{i=1}^{N} \frac{cov(f_1(x_i), f_2(x_i))}{\sigma_{f_1(x_i)} \sigma_{f_2(x_i)}}
\]
where $N$ is the number of test samples, $cov(f_1(x_i), f_2(x_i))$ is the covariance between two network output vectors for input $x_i$, and $\sigma_{f(x_i)}$ is the standard deviation of the output vector.

Table 6 compares the average correlations of Prune and Tune Ensembles with that of a dense bootstrap aggregated ensemble on CIFAR-10 with ResNet-18 and an ensemble size of eight models. We report results for both random and anti-random prune and tune ensembles as well as an anti-random ensemble tuned with bootstrap aggregated data. Bootstrap aggregation involved training on a random 90\% subset of the training data for each child member.

\begin{figure}[t]

\begin{tabularx}{\columnwidth}{X c c c}
\toprule
~ & Acc. $\uparrow$ & Corr. $\downarrow$ & Std. Dev. $\downarrow$ \\
\midrule
PAT (R) & 0.840 & 0.8787 & $\pm$ 0.0449 \\
PAT (AR) & 0.844 & 0.8679 & $\pm$ 0.0410 \\
PAT (AR + BAG) & 0.839 & 0.8501 & $\pm$ 0.0509 \\
Dense (BAG) & 0.834 & 0.8621 & $\pm$ 0.0476 \\
\bottomrule
\\
\includegraphics[width=\columnwidth]{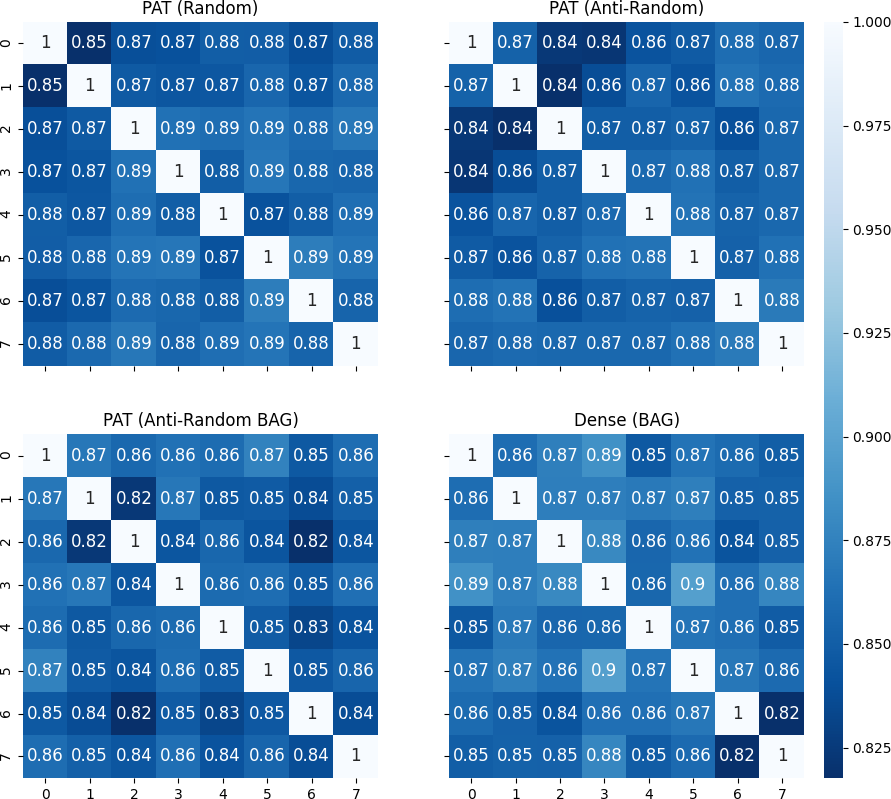}

\end{tabularx}
\captionof{table}{Pairwise correlation values for ResNet-18 on CIFAR-10. (R/AR) denote random/anti-random pruning. (BAG) denotes bootstrap aggregation.}

\end{figure}

The pairwise correlations are very similar to each other for a comparable accuracy. Prune and Tune ensembles with bagging has the lowest correlation of all methods. Notably, bagging tends to result in worse accuracy due to the fewer training samples it sees. We do see a significant difference in diversity between both random and anti-random pruning methodologies.

\subsubsection{Prediction Disagreement Ratio}

Prediction disagreement is the ratio of test data that two models disagree on. We compute this in a pairwise fashion for each sample in the test set and we average the total scores together for each model.

\[
d_{dis}(f_1, f_2) = \frac{1}{N} \sum_{i=1}^N argmax \ (f_1(x_i)) \neq argmax \ (f_2(x_i))
\]

We compare our results using predictions from the large budget experiment for CIFAR-10 using WideResNet-28-10 as our model. We find comparable results for our prediction disagreement ratio with the best performing ensemble methods. The diversity of Prune and Tune Ensembles is competitive with that of Dense Ensembles trained for 4x longer.

\subsubsection{Kullback-Leibler Divergence}

Also known as relative entropy, KL-Divergence approximately measures how different one probability distribution is from another:

\[
d_{KL}(f_1, f_2) = \frac{1}{N} \sum_{i=1}^N f_1(x_i) \log \left( \frac{f_1(x_i)}{f_2(x_i)} \right)
\]

We use the same predictions as for the large budget experiment for CIFAR-10 using WideResNet-28-10. We find that the average KL divergence of prune and tune ensembles is greater than that of reported results for other ensemble methods.

\begin{table}[t]
\begin{tabularx}{\columnwidth}{X l l}
\toprule
Methods & $d_{dis}$ $\uparrow$ & $d_{KL}$ $\uparrow$ \\
\midrule
Treenet$^*$ & 0.010 & 0.010 \\
BatchEnsemble$^*$ & 0.014 & 0.020 \\
LTR Ensemble$^*$ & 0.026 & 0.057 \\
EDST Ensemble$^*$ & 0.026 & 0.057 \\
PAT Ensemble & 0.036 & 0.090 \\
\midrule
MIMO$^*$ & 0.032 & 0.081 \\
Dense Ensemble$^*$ & 0.032 & 0.086 \\
\bottomrule
\end{tabularx}
\caption{Disagreement ratio and KL divergence between ensemble members on CIFAR-10 with WideResNet-28x10. Methods marked with $^*$ are results reported from \cite{liu2021freetickets}.}
\end{table}

\subsubsection{Visualizing The Diversity of Child Networks in Representation Space}

Pruned networks derived from the same parent can result in significantly different representations of the feature space. We use t-distributed stochastic neighbor embedding (t-SNE) to visualize the representation of test data in lower dimensional clusters \cite{JMLR:v9:vandermaaten08a}.

T-SNE converts similarities between data points to joint probabilities and tries to minimize the Kullback-Leibler divergence between the joint probabilities of the low-dimensional embedding $q_{ij}$ and the high-dimensional embeddings $p_{ij}$.

First, conditional probabilities are assigned to pairs of objects according to gaussian kernels.

\[
p_{j|i} = \frac{exp(- ||x_i - x_j||^2 / 2 \sigma_i^2)}{\sum_{k \neq i} exp(- || x_i - x_k ||^2 / 2 \sigma_i^2)}
\]

where, $x_i$ and $x_j$ are input samples. The joint probability is then defined as:
 
\[
p_{ij} = \frac{p_{j|i} + p_{i|j}}{2N}
\]

The low dimensional probabilities $q_{ij}$ are defined according to the means of a student-t distribution.

\[
q_{ij} = \frac{(1 + ||y_i - y_j||^2)^{-1}}{\sum_k \sum_{l \neq k} (1 + ||y_k - y_l ||^2)^{-1}}
\]

The locations of the points $y$ are determined by minimizing the Kullback-Leibler divergence of two distributions P and Q using gradient descent.

We perform this algorithm for a trained parent and two randomly pruned child networks on CIFAR-10 with the LeNet-L model. Each child is tuned for only one epoch using ADAM and a learning rate of $\eta=0.001$. We use a fixed random seed for t-SNE and we plot the results of model representations, predicted classes, and mispredicted classes. 

Figure 4 displays significant diversity in the representations of the feature space as a result of topological differences in these networks, despite the fact that both children are derived from an identical parent and only tuned for one epoch.

\begin{figure}[t]
\centering
    {\includegraphics[width=\columnwidth]{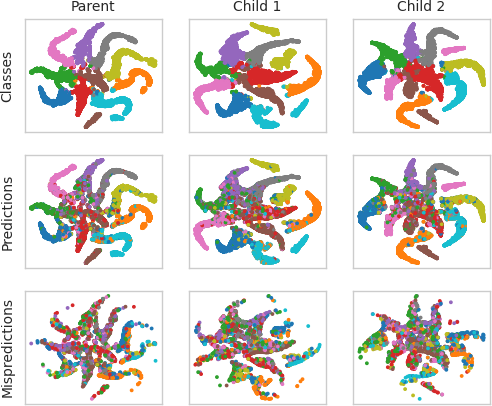}}
    \caption{T-SNE cluster representations of CIFAR-10 and LeNet. The first row is the representation of classes in CIFAR-10 that each our models produces. The second row displays the results of our model predictions and the final row displays the incorrect predictions that our models made.}
\end{figure}

\subsection{Visualizing the Loss Landscapes of Pruned Child Networks}

We use the repository \cite{li2018visualizing}, to explore the loss landscapes of neural networks as they undergo both pruning and tuning.

We train a ResNet-56 model for 300 epochs using Stochastic Gradient Descent with Nesterov momentum. The learning rate is $\eta_1=0.1$ and decays to $\eta_2=0.01$ at 150 epochs, $\eta_3=0.001$ at 225 epochs and $\eta_4=0.0001$ at 275 epochs. A batch size of 128 is used along with a weight decay $\gamma=0.0005$. We prune the network by eliminating a random 50\% of the weights and we tune the child network for three additional epochs with a constant learning rate $\eta=0.001$.

Countour plots are created by choosing a center point $\theta$ and choosing two orthogonal directional vectors $\delta$ and $\eta$. The loss landscape is then plotted by linearly interpolating over these two directional vectors.

\[
f(\alpha, \beta) = L(\theta + \alpha \delta + \beta \eta)
\]

The authors get around the scale variance problem by using filter wise normalization of the direction vectors. A network can have widely varying weight magnitudes between layers and different sensitivities to perturbative noise. To alleviate this, the directional vectors are scaled according to the norms at a filter level. For a network with parameters $\theta$, a gaussian vector $d$ is generated with the same dimensions of $\theta$. Then, for a given filter $d$:

\[
d_{i,j} = \frac{d_{i,j}}{|| d_{i,j}||} || \theta_{i,j} ||
\]

where $d_{i,j}$ represents the $jth$ filter of the $ith$ layer of $d$, and $|| . ||$ denotes the Frobenius norm.

Figures 5, 6, and 7 display the results of this visualization. We note the apparent simplification of the loss landscapes of pruned and tuned networks. The optima for children are not as low as the parent, but they do settle in areas of the landscape that are shallow and wide, suggesting good locations for generalization.

\begin{figure}[h!]
    \includegraphics[width=\columnwidth]{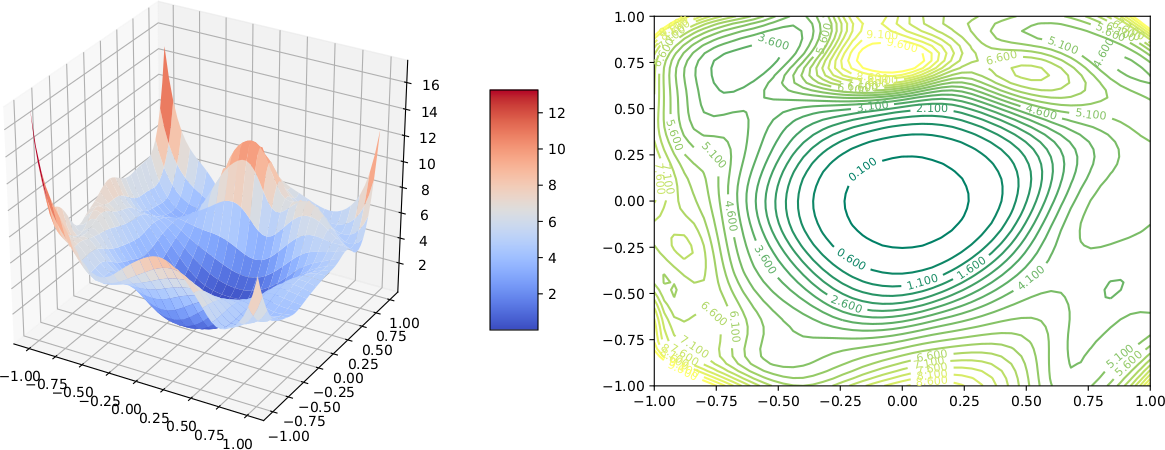}
    \caption{The optimized landscape of a trained ResNet-56 model.}
    \vspace{0.5in}
    
    \includegraphics[width=\columnwidth]{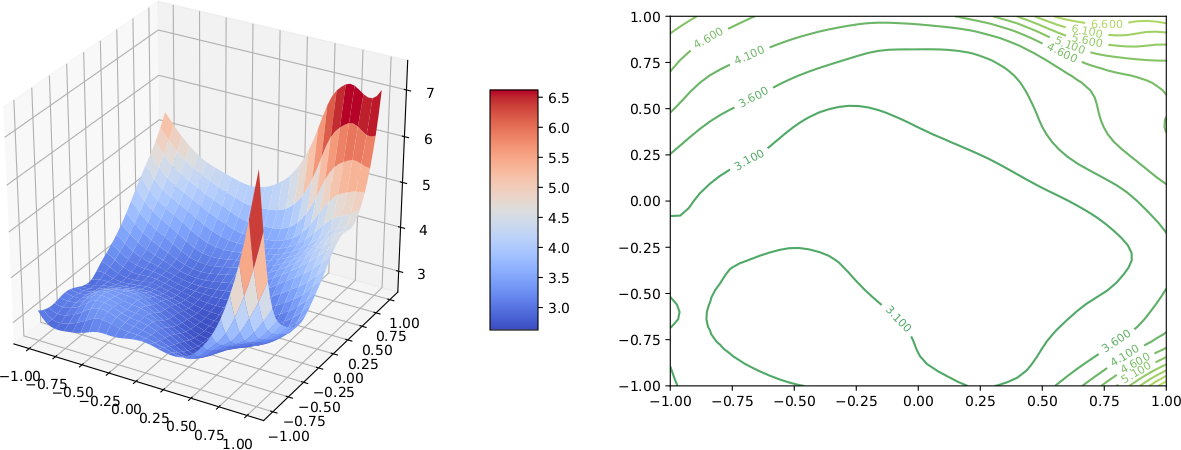}
    \caption{The landscape of a trained ResNet-56 model with 50\% of its parameters pruned.}
    \vspace{0.5in}

    \includegraphics[width=\columnwidth]{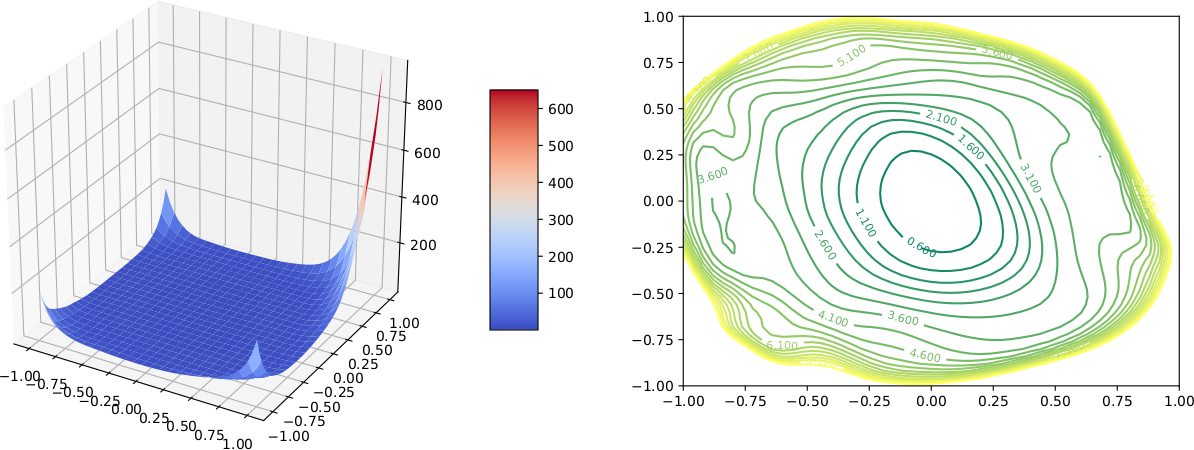}
    \caption{The landscape of a pruned ResNet-56 model after three additional epochs of fine tuning.}
    \vspace{0.5in}
\end{figure}

\end{document}